\documentclass[conference]{IEEEtran}
\IEEEoverridecommandlockouts

\usepackage{amsmath,amssymb,amsfonts}
\usepackage{algorithmic}
\usepackage{graphicx}
\usepackage{booktabs} 
\usepackage{textcomp}
\usepackage{xcolor}
\usepackage{verbatim}
\def\BibTeX{{\rm B\kern-.05em{\sc i\kern-.025em b}\kern-.08em
    T\kern-.1667em\lower.7ex\hbox{E}\kern-.125emX}}

\usepackage{float}

\usepackage{bibunits}
\usepackage{algorithm}
\usepackage{caption}
\usepackage[hidelinks]{hyperref}

\usepackage{tabularx}
\newcolumntype{Y}{>{\centering\arraybackslash}X}
\usepackage{booktabs}
\usepackage{multirow}
\usepackage{arydshln}
\usepackage{subcaption}

\usepackage{comment}

\begin{document}

\title{LTOS: Layout-controllable Text-object Synthesis
via Adaptive Cross-attention Fusions
}

\author{\IEEEauthorblockN{1\textsuperscript{st} Xiaoran Zhao}
\IEEEauthorblockA{\textit{National University of Defense Technology} \\
}
\and
\IEEEauthorblockN{2\textsuperscript{nd} Tianhao Wu}
\IEEEauthorblockA{\textit{Nanyang Technological University} \\
}
\and
\IEEEauthorblockN{3\textsuperscript{rd}Yu Lai}
\IEEEauthorblockA{\textit{National University of Defense Technology} \\
}
\and
\IEEEauthorblockN{4\textsuperscript{th} Zhiliang Tian$^\ast$}
\IEEEauthorblockA{\textit{National University of Defense Technology} \\
}
\and
\IEEEauthorblockN{5\textsuperscript{th} Zhen Huang$^\ast$\thanks{$^\ast$ Corresponding Author }}
\IEEEauthorblockA{\textit{National University of Defense Technology} \\
}
\and
\IEEEauthorblockN{6\textsuperscript{th} Yahui Liu}
\IEEEauthorblockA{\textit{Huawei Technologies Ltd} \\
}
\and
\IEEEauthorblockN{7\textsuperscript{th} Zejiang He}
\IEEEauthorblockA{\textit{National University of Defense Technology} \\
}
\and
\IEEEauthorblockN{8\textsuperscript{th} Dongsheng Li}
\IEEEauthorblockA{\textit{National University of Defense Technology} \\
}
}

\maketitle

\begin{abstract}
Controllable text-to-image generation synthesizes visual text and objects in images with certain conditions. However, existing visual text rendering and layout-to-image generation tasks focus on single modality generation or rendering, leaving yet-to-be-bridged gaps between the approaches correspondingly designed for each of the tasks. In this paper, we introduce a unified task called layout-controllable text-object synthesis (LTOS), which merges text rendering and layout-based object generation. We construct a layout-aware text-object synthesis dataset for the LTOS task, containing well-aligned labels visual text, and object information. We also propose a layout-controllable text-object adaptive fusion (TOF) model, which generates images with legible visual text and plausible objects. Specifically, we construct a visual-text rendering to synthesize text and employ an object-layout control module to generate objects while integrating the two modules to harmoniously generate and integrate text and objects in images. To enhance the image-text integration, we propose a self-adaptive cross-attention fusion that forces the image generation to attend more to important text spans. Within such a fusion, we use a self-adaptive learnable factor to learn to control the influence of cross-attention outputs. Experiments show that our model excels the strong baselines and obtain SOTA in LTOS. We release our dataset and codes \footnote{https://anonymous.4open.science/r/TOF-D5EE}.
\end{abstract}


\section{Introduction}
Controllable text-to-image (T2I) generation \cite{cao2024controllable} synthesizes text contents and objects on an image, which makes the generated images harmoniously juxtaposed with visual text \cite{auvil2007flexible} and has many applications including emoji crafting \cite{suryadevara2019emojify} and poster design \cite{huo2022study}. Text rendering and layout-to-image, as two popular tasks, have appealed to extensive research interests. (1) \textbf{Text rendering} in images refers to generating clear and legible text in images, which needs to be visually compatible with the image in terms of texture and depth.
Latest methods \cite{zhao2023udifftext, tuo2023anytext, yang2024glyphcontrol}  typically encode the text input into an image with the backbone of LDMs \cite{rombach2022high} or ControlNet  \cite{zhang2023adding} to achieve high-quality text rendering.
(2) \textbf{Layout-to-image generation} task is to generate an image according to a given layout map such as bounding boxes with object categories \cite{li2022grounded} or a semantic segmentation map \cite{he2023localized, lin2014microsoft}. 
The core of this task is to accurately control the location of generated objects, which can be seen as the reverse task of object detection \cite{zou2023object}. 
\begin{figure}[h]
  \centering
  \includegraphics[width=\linewidth]{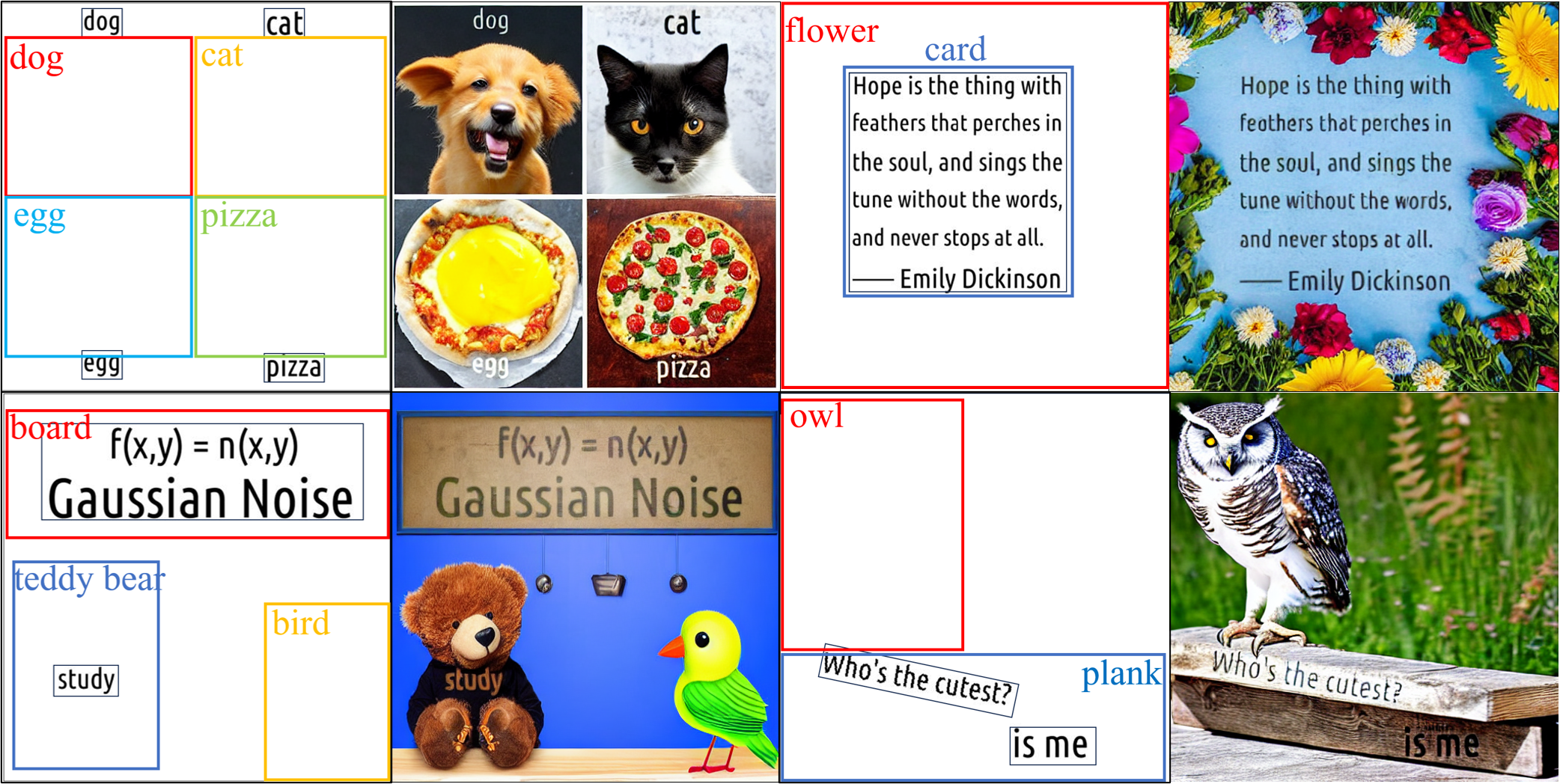}
  \caption{Our model TOF's examples. Given predefined visual text and object information, TOF generates objects following the layout map while rendering text consistent with the image's depth, texture, and geometry.}
  \label{fig: teaser}
\end{figure}
Despite the booms on the two tasks, the majority of these approaches can only accurately control one singular modality: 
text rendering caters more to the quality of the generated visual text and controls objects simply with prompts, being hard to define the positions of the objects; layout-to-image generation focuses on controlling objects, ignoring the clarity and rationale of the generated visual text. 
Such differentiation in focus led to a significant and yet-to-be-bridged gap between the design of methods to control text and objects. 

To mitigate this challenge, we integrate the aforementioned two mainstream tasks - text rendering and layout-to-image generation - into a singular task: layout-controllable text-object synthesis (LTOS). 
LTOS's objective is to generate an image with visual text, where the location of the objects and text can be accurately controlled by providing a layout map together with object categories and visual text information.
Due to the lack of datasets containing both textual and object layout information for this task, we develop a layout-aware text-object synthesis (LTOS) dataset. It contains well-aligned multi-modal label information including image captions (prompts), word-level visual text, and object bounding boxes with category labels.
LTOS dataset is easily expanded with a three-step workflow and has high quality comparable to the real dataset.
 
Based on our dataset, we propose a layout-controllable text-object adaptive fusion (TOF) model that synthesizes images with high-quality object and visual text conditioned on the given object layouts and text contents. TOF consists of 1) visual-text rendering module, 2) an object-layout control module, and 3) a text-object self-adaptive fusion module. 
Our \textit{visual text rendering module} customizes the instructions for rendering text layouts with multi-region and multi-direction. Given the spatial layout information of objects (i.e. object categories with matching bounding boxes), our \textit{object-layout control module} generates images with each object placed at the predetermined location. 
To harmoniously integrate text content and image objects, we integrate the above two modules to jointly control the layout of visual text and objects.
To balance visual-text rendering and object-layout control on the integration, we propose a \textit{text-object self-adaptive fusion module} to select important text information for image generation via a cross-attention mechanism \cite{vaswani2017attention}. To absorb the cross-attention outputs, we employ a self-adaptive learnable factor that learns to fetch the influence of cross-attention outputs for image generation. 
Experimental results show that our model excels the strong baselines and obtain SOTA on our dataset, indicating the effectiveness of our design. 
Our contributions are: (1) We merge text rendering and layout-to-image generation tasks and define a unified task: layout-controllable text-object synthesis (LTOS) to synthesize text and objects into an image. (2) We construct a layout-aware text-object synthesis dataset for LTOS. (3) We propose a layout-controllable text-object adaptive fusion for LTOS, customizing the text layout and controls text-object synthesis.

\section{Method}

\begin{figure}[h]
  \centering
  \includegraphics[width=\linewidth]{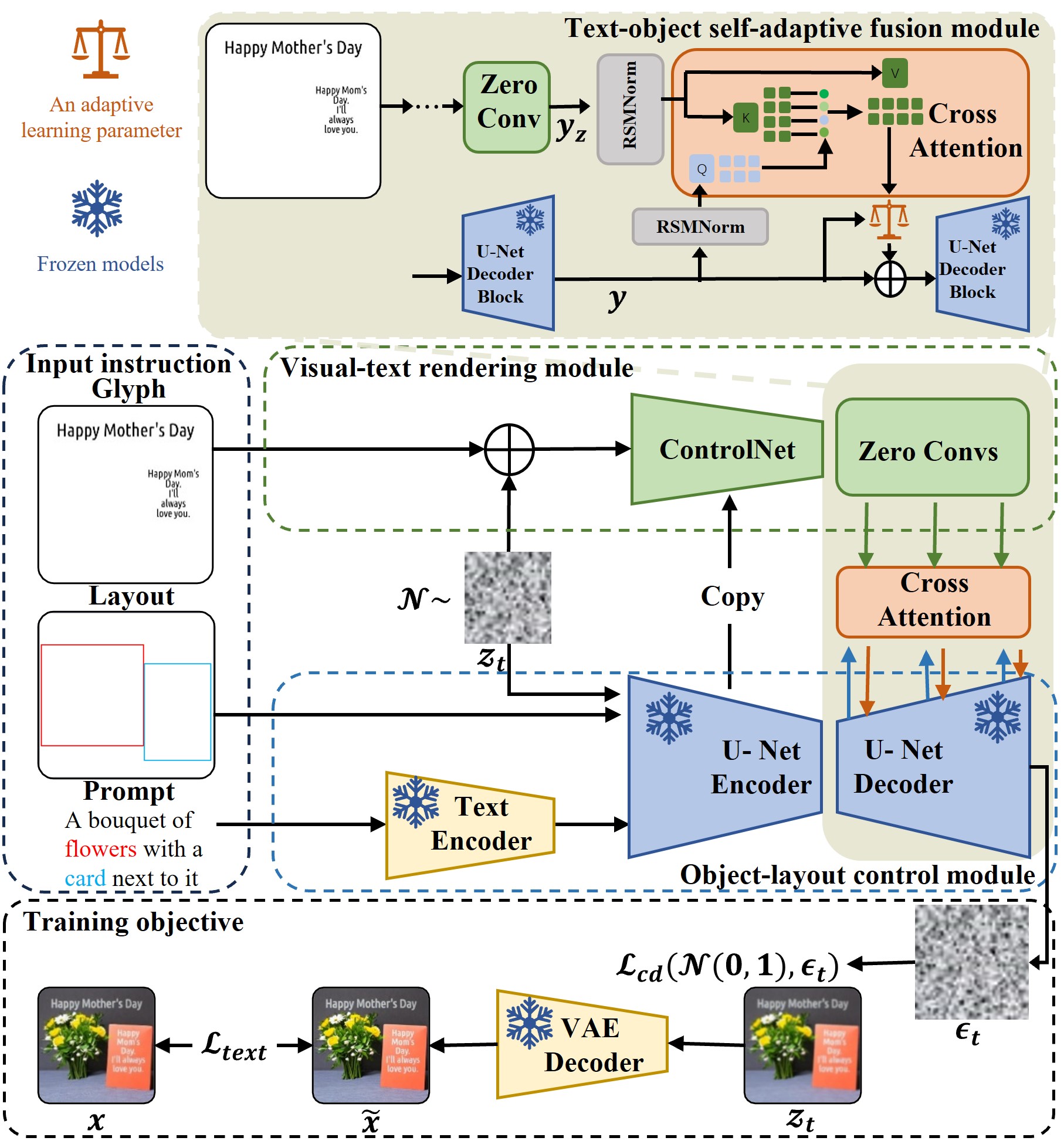}
  \caption{TOF framework. Given the input with a prompt, a glyph image, and a layout map, TOF achieves controllable text-object synthesis. 
  }
  \label{fig: pipeline}
\end{figure}

As Fig.~\ref{fig: pipeline}, our model consists of a visual-text rendering module and an object-layout control module for text-object integration control (Sec.~\ref{sec: integration control model}). Further, we propose a text-object self-adaptive fusion to balance text and object (Sec.~\ref{sec: self-adaptive fusion module}). 

\subsection{Task formulation}
\label{sec: task formulation}
For our LTOS task, we denote LTOS's input $I$ consists of a text prompt describing the image content $\mathbf{c}$, a layout map $O$ consisting of object categories (with its bounding boxes), a glyph image $G$ reflecting text content, position, and shape, $I = \langle \mathbf{c}, \mathrm{O}, \mathrm{G} \rangle$. In an input with image carrying $n$ objects and $m$ text regions $I= \langle \mathbf{c}, \{(o_1, b_{o_1}), \ldots, (o_n, b_{o_n})\}, \{(t_1, g_{t_1}), \ldots, \\(t_n, g_{t_m})\} \rangle $, 
where $o_i$ and $t_j$ denote the object category and the visual text content respectively, $b_{o_i}$ is the bounding box for $o_i$ and $g_{t_j}$ is the corresponding glyph region for $t_j$ ($1 \leq i \leq n, 1 \leq j \leq m$).
Given the input, our target is generating an image satisfying 1) the generated objects are within each bounding box, consistent with the specified categories and the text descriptions; 2) the visual text formatting (position, font size, and distortion) follows the exact guidance of the glyph image; and 3) the rendered text isseamlessly integrated with the generated image in depth, texture, and geometry. 

\subsection{Dataset Construction}
Existing datasets with both visual-text and it aligned object layouts with object category annotation aren't available. Besides, datasets used for text rendering face some limitations: 1) Each text area is large than 10\% of the entire image (almost no tiny-font text) \cite{chen2024textdiffuser,tuo2023anytext}; 2) visual text appears in center areas of the image and rarely in border areas \cite{ma2023glyphdraw}.
Hence, it leads to potential issues: the incapability to generate tiny-font text and high-quality visual text around border areas.

Considering the limitations, we propose a layout-aware text-object synthesis dataset, a multi-modal dataset with comprehensive well-aligned labels in both visual text and object. 
We choose the Flickr30K \cite{flickr30k} dataset as the basis for embedding textual information at appropriate places within images, and the Flickr30K Entities dataset \cite{flickrentitiesijcv} to supplement object category annotated bounding boxes. 
However, Flickr30K doesn't include visual text information, which is crucial for our task.
Fortunately, SynthText \cite{gupta2016synthetic} provides a potential solution by enriching the data with coarse visual text rendering.

The dataset construction involves three steps. \textbf{Step 1}: Select regions suitable for text rendering by using texture and depth information of images; \textbf{Step 2}: Set filtering rules to generate text as required and color the text based on Color-model \cite{gupta2016synthetic};
\textbf{Step 3}: Perform Poisson image editing \cite{perez2023poisson} to blend the text into the scene.
The filtering rules in step 2 include: 1) Text region number ranges from 1 to 8; 2) The number of text lines ranges from 1 to 8; 3) The text is generated in random regions that is suitable for image’s text rendering; 4) Each text region exceeds 2\% of the image in length and width, with a minimum area of 5\% of the entire image; 5) Text is randomly rotated, twisted, bolded, and bordered.

Following that workflow, we notice that the background of text regions recognized as suitable for text rendering was too cluttered. 
To address the issues, we first replaced the original depth maps via recent Zoedepth \cite{bhat2023zoedepth} to estimate more precise depth; then, we applied PPOCRv3 \cite{li2022pp} to filter out all the images whose textual content is not satisfactory, i.e. containing low-quality renderings. These above designing enables stable and higher-quality text rendering, comparable to the real dataset in terms of depth, texture, and geometry.

\begin{figure}[!t]
  \includegraphics[width=\linewidth]{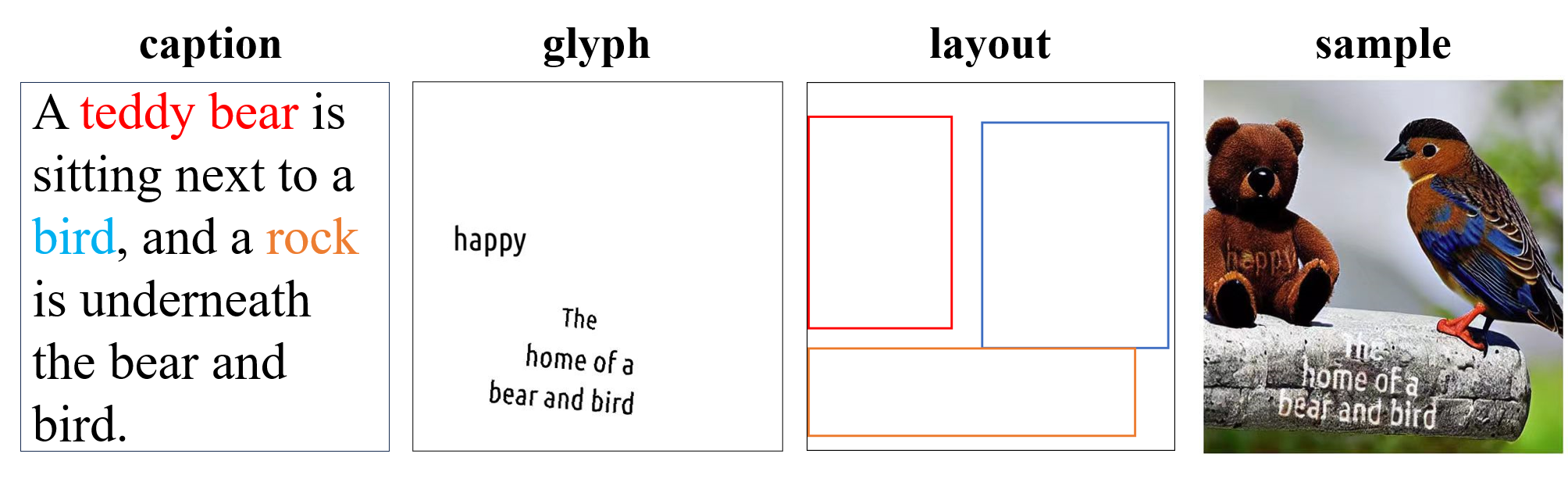}
  \caption{An example sample of our LTOS dataset.
  }
  \label{fig: dataset}
\end{figure}

LTOS dataset includes 228k samples. Each sample has: 1) an image with rendered text, 2) a caption of the image used as the prompt, 3) grounded word-level text annotations with enough information to 
obtain the corresponding glyph image, and 4) grounded object annotations. LTOS has some fine-grained attributes: depth of the visual text, multiple thicknesses, multiple text regions, and different rotation angles. As shown in the Fig ~\ref{fig: dataset}, an example of our LTOS dataset.

\subsection{Text-object integration control model}
\label{sec: integration control model}
To handle LTOS task, we propose a visual-text rendering module and an object-layout control module to render text on the image and generate images with objects conditioned on specific layouts.
(1) \textbf{visual-text rendering module}
generates visual text with multi-regional and multi-directional layouts and injects the feature of glyph images as conditions to guide the text-rendered images.
Inspired by GlyphControl \cite{yang2024glyphcontrol} and AnyText \cite{tuo2023anytext}, we employ a trainable ControlNet \cite{zhang2023adding} to achieve text rendering with controlling the geometric structures accurately. But this module hardly controls the accurate layout of the generated objects, especially when faced with multiple objects.
(2) To control the generation of objects more precisely, we propose a \textbf{object-layout control module}
extracts features from the object layout map $\mathrm{O}$ based on GLIGEN\cite{li2023gligen}, and generate images with the layout. The above mechanism achieves controllable visual-text rendering and object generation by connecting  visual-text rendering and object-layout control. 

\subsection{Text-object self-adaptive fusion module}
\label{sec: self-adaptive fusion module}
To better integrate text and objects,
we propose a text-object self-adaptive fusion that adaptively bridges the visual-text rendering and object-layout control.
The motivation is that the aforementioned module (Sec. \ref{sec: integration control model}) encounters two issues:
(1) it is difficult to balance text rendering and object generation. Over-training text rendering branch leads to the distortion of some generated objects while under-training text rendering makes it hard to render clear and legible text.
(2) rendering multiple text areas on diverse objects results in blurred and unclear visual text in images.
To mitigate this issue, we propose to bridge text and object via adaptive cross-attention, where (1) the cross-attention fetches textual information to help image generations and (2) a self-adaptive learnable factor adaptively learns to leverage the information from the cross-attention.
The cross-attention consists of three sub-modules.
\textbf{Cross-attention backbone} follows the vanilla cross-attention \cite{2017Attention}. 
As shown in Fig.~\ref{fig: pipeline}, each block $B$ in the object-control module corresponds to a trainable copy block with a zero-convolution layer, together denoted as $B_z$, in the visual-text module.
we inject cross-attention layers between the output of $B$ and  $B_z$, denoted as $\mathbf{y}$ and $\mathbf{y}_z$ respectively.
    The cross-attention layer takes $\mathbf{y}$ and $\mathbf{y}_z$ as input and outputs a feature vector $\mathbf{y}_a$ that combines the visual text and object information.
\textbf{RMSNorm operation} enables the rapid convergence \cite{zhang2019root} in our cross-attention backbone. We applied RMSNorm to the input of cross-attention $\mathbf{y}$ and $\mathbf{y}_z$ (i.e. the output of object-control and zero-convolution). The output of RMSNorm serves as the input of cross-attention:
$\mathbf{y}_a = \text{cross-attention}(\text{RMSNorm}(\mathbf{y}),\text{RMSNorm}(\mathbf{y}_z))$.
\textbf{Self-adaptive learnable factor} learns to decide how much information from cross-attention is used (i.e. influence of cross-attention's output $\mathbf{y}_a$). 
It allows the model to adjust the weights of additional visual text $\mathbf{y}_a$ and object $\mathbf{y}$. We apply $\text{tanh}(\cdot)$ to the learnable factor $\alpha$ to constrain  $\alpha$'s range. To linearly weighted integrate $\mathbf{y}_a$ and $\mathbf{y}$, we element-wise sum up the object $\mathbf{y}$ and $\mathbf{y}_a$ adjusted by the factor $\alpha$. Then, we concatenate the summation with $\mathbf{y}$ \footnote{Concatenating $\mathbf{y}$ and $\mathbf{y}_a$ instead of summing generates distorted objects.} as the output of this module $\mathbf{y}_f$, as 
$\mathbf{y}_f = \text{concat}(\mathbf{y}, \mathbf{y} + \text{tanh}(\alpha) * \mathbf{y}_a)$.
    

\subsection{Training objective}
\label{sec: training objective}
The training is as follows: we feed the given input image $x\in\mathbb{R}^{H \times W \times 3}$ in a pre-trained VAE \cite{kingma2013auto} to get its latent vector $\mathbf{z}_0\in\mathbb{R}^{h\times{w}\times c}$.
In the object-layout control, 
each gated self-attention layer's input is denoted as $\mathbf{z}_{b}$. 
In the visual-text rendering, 
we extract its feature as $\mathbf{z}_g$\cite{bansal2023universal, kim2023diffblender}.
During the diffusion of training, we sample a time $t \sim \text{Uniform}(0, 1000)$, a Gaussian noise $\epsilon \sim \mathcal{N}(0, 1)$, and corrupt $\mathbf{z}_0$ to generate noisy latent images $\mathbf{z}_t$.
Our training objective consists of: (1) controllable diffusion loss $\mathcal{L}_{cd}$, where we use a network \(\epsilon_\theta\) to estimate the noise that is  used in the latent image $\mathbf{z}_t$ \cite{ho2020denoising}  as:
\begin{equation}
    \mathcal{L}_{cd} =\mathbf{E}_{\mathbf{z}_0,\mathbf{z}_b, \mathbf{z}_g, \mathbf{t}, \epsilon \sim \mathcal{N}(0, 1)} [ \left\| \epsilon - \epsilon_{\theta}(\mathbf{z}_{t}, \mathbf{z}_{b}, \mathbf{z}_g, \mathbf{t}) \right\|^2_2]
\end{equation}
(2) Text perceptual loss $\mathcal{L}_{text}$ \cite{tuo2023anytext}, where we minimize the gap between rendered text in generated image $\tilde{x}$ and that in the ground truth image $x$. 
Specifically, with our grounded word-level annotations, we crop each pair of $x$ and $\tilde{x}$ into sub-images $\mathcal{S}=\{s_1,...,s_j\}$ and $\mathcal{S}'=\{s_1',...,s_j'\}$ with one text region in each $s_*$. 
We feed $\mathcal{S}$ and $\mathcal{S}'$ into PP-OCRv3 \cite{li2022pp} and get the mean squared error  \cite{chai2014root} (MSE) between the features $f_i$ and $f_i'$:
 $\mathcal{L}_{text} = \sum_{i}\frac{\phi(t)}{hw} \sum_{h,w}\|f_i - f_i'\|^2_2 \ $
, where $f_i$ and $f_i' \in \mathbb{R}^{h \times w \times c}$ is output of the last fully connected layer of $x$ and $\tilde{x}$; $\phi(t)$ is a weight adjustment \cite{tuo2023anytext} same as the coefficient of diffusion in \cite{ho2020denoising}. 
The overall training objective is $\mathcal{L}= \mathcal{L}_{cd} + \lambda * \mathcal{L}_{text}$, where $\lambda$ balances $\mathcal{L}_{cd}$ and $\mathcal{L}_{text}$.


\section{Experiment}
\noindent\textbf{Settings.}\label{Experiments settings}
Our backbone is based on GLIGEN \cite{li2023gligen}. We initialize the visual-text rendering with pre-trained ``bounding box + text'' weights.   
Our dataset has 217,820 training and 11,113 testing samples.
For diffusion, we set the downsampling factor $f=8$ and the latent dimension is 64$\times$64$\times$4. We train the model for 1.3 M iterations with batch size 24. Since LTOS is a new task, we compare with baselines on related tasks: 
(1) layout-to-image generation tasks, GLIGEN \cite{li2023gligen}, (2) text rendering tasks, TextDiffuser \cite{zhang2024brush}, AnyText \cite{tuo2023anytext}, and GlyphControl \cite{yang2024glyphcontrol} (SOTA on MARIO-10M, AnyWord-3M, LAION-Glyph datasets, respectively).
We use three metrics: 1) OCR. Acc \cite{li2022pp}, the word-level accuracy of rendered texts; 2) Normalized Edit Distance (NED) \cite{Marzal1993pami},the similarity between two strings;
3) AP \cite{li2021image}, the accuracy of object generation.
 
\noindent{\textbf{Main results.}} As Tb~\ref{tab: main}, our model significantly outperform all state-of-the-art baselines on the benchmarks in all metrics (+7.10\% on OCR. ACC and +6.63\% on NED). TextDiffuser generates text layout which is not effective at rendering non-center text.
When AnyText renders multiple texts with different fonts, sizes and rotation angles on backgrounds with different textures, text's clarity is affected.
Notice that GlyphControl performs poorly when rendering rotated text beyond 20 degrees especially on complex background textures. Even after fine-tuning on LOST dataset, results are not ideal.



\noindent{\textbf{Ablation studies.}}
In Tb~\ref{tab: Ablations}, removing any of the modules results in worse results. Row 2 and 5 verify the effectiveness of \textit{text-object self-adaptive fusion}. Row 3 verify the \textit{self-adaptive mechanism} in Sec.~\ref{sec: self-adaptive fusion module}.
Row 4 in Tb~\ref{tab: Ablations} suggests that  \textit{text perceptual loss} enhances the quality of text generation. 

\noindent{\textbf{Analyses on attention-layers.}} We test the model with different numbers and settings of injecting cross-attention layers (4-layer Dense: four layers after the 0-th to 3-rd zero-convolution layers; 4-layer Sparse: four layers after the 0-th, 3-rd, 6-th and 9-th zero-convolution layers) in Tb~\ref{tab: main}. It suggests that increasing the number of layers doesn't necessarily lead to better results. Instead, the over-introduction of layers leads to significant results degradation. We use the 4-layer Dense setting in this paper, which leads to the best results.

\begin{table}[tb!]
  \caption{TOF compared to baselines and analysis on cross-attention. * means finetuning on LTOS dataset. - indicates metrics do not fit baselines considering the baselines' original tasks.}
  \label{tab: main}
  \resizebox{\linewidth}{!}{
  \begin{tabular}{@{}cccc@{}}
    \toprule
    \textbf{Method} & \textbf{OCR. ACC} $\uparrow$ & \textbf{NED} $\uparrow$ & \textbf{AP} $\uparrow$\\
    \midrule
    TextDiffuser \cite{chen2024textdiffuser} & 0.2551& 0.4824&-\\
    $\text{GlyphControl}^*$ \cite{yang2024glyphcontrol} & 0.3643& 0.5662& -\\
    AnyText \cite{tuo2023anytext} & 0.2862&0.5074&-\\
    GLIGEN \cite{li2023gligen} & -& -& 0.4958 \\
    Our TOF&\textbf{0.4353}&\textbf{0.6325}&\textbf{0.5658}\\
   \toprule
    4-layer Sparse & 0.4039& 0.5956&0.4791\\
    2-layer Dense & 0.4203&0.6161&0.4571\\ 
    7-layer Dense & 0.4115&0.6015&0.4671\\
    10-layer Dense & 0.3533&0.4402&0.4710\\
    4-layer Dense (Our TOF) &  \textbf{0.4353}&\textbf{0.6325}&\textbf{0.5658}\\
  \bottomrule
\end{tabular}}
\end{table}

\begin{figure*}[!t]
  \includegraphics[width=\textwidth]{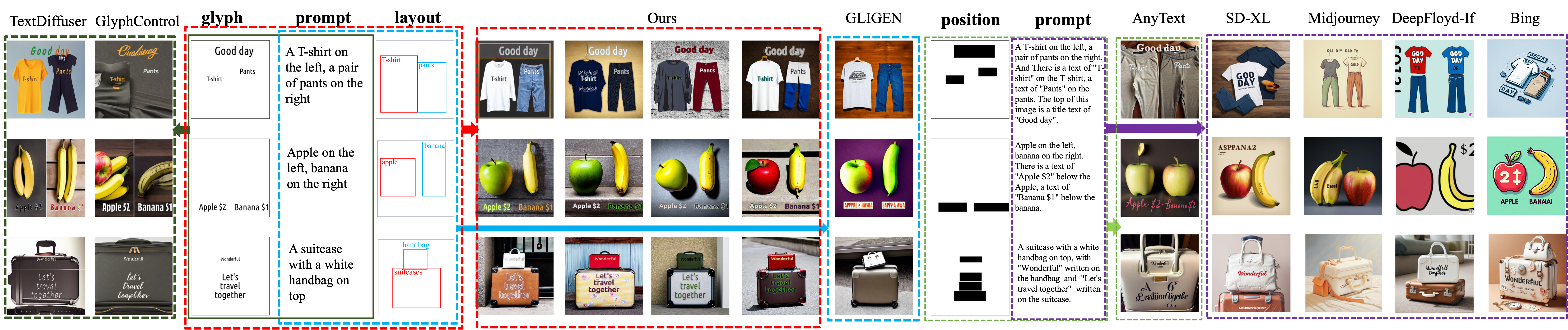}
  \caption{Qualitative comparison of TOF with the baselines and widely used T2I generation APIs.
  }
  \label{fig: goodcase}
\end{figure*}

\begin{figure*}[!t]
  \includegraphics[width=\textwidth]{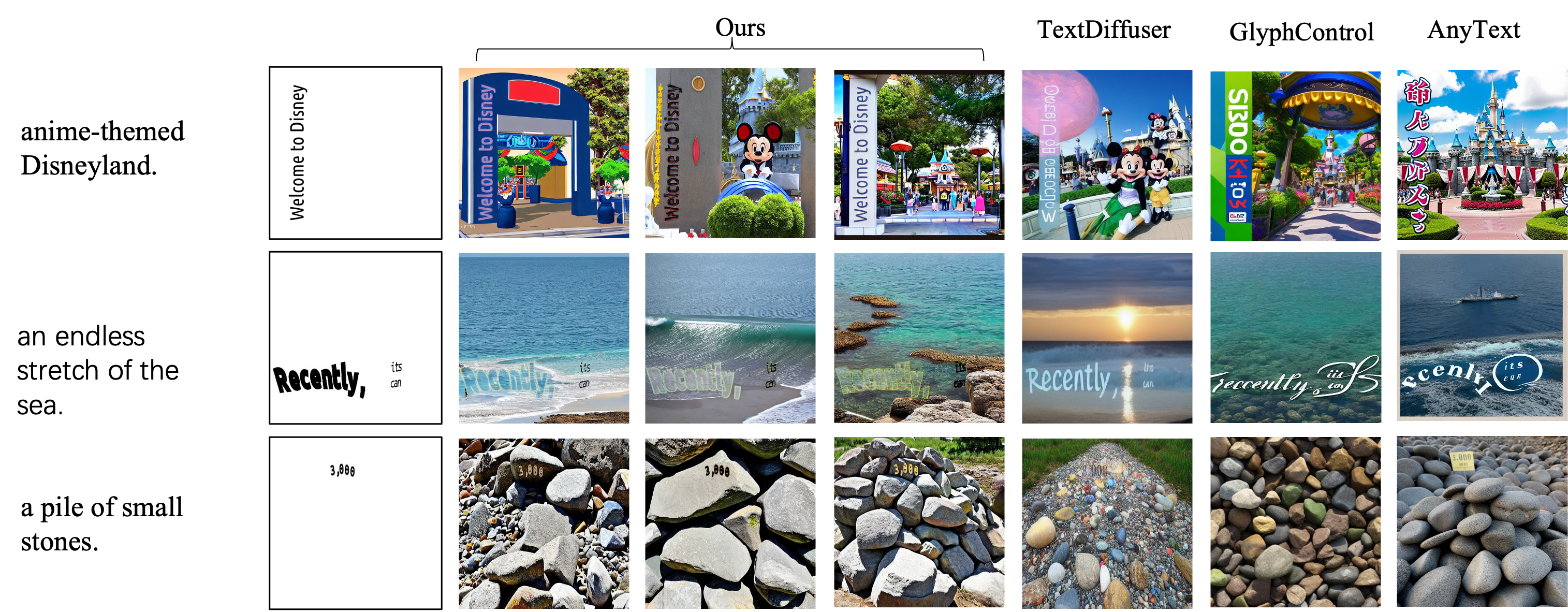}
  \caption{Qualitative comparison of TOF and the baselines. 
  Due to the differences between our input and baselines, we keep the input to be the same based on the input format accepted by all models. 
  }
  \label{fig: special}
\end{figure*}

\noindent{\textbf{Qualitative results.}} 
Our qualitative results compared with baselines and widely used T2I generation APIs (SD-XL, Midjourney, DeepFloyd IF, and Bing Image Creator) are visualized in Fig.~\ref{fig: goodcase}.
It shows that our method renders more legible, clearer visual texts and plausible objects at the predefined positions. Our model supports arbitrary rotation of text area, tiny font, and different thicknesses of the font in Fig.~\ref{fig: special}.
We observed limitations in baselines.
AnyText can't easily wrap and rotate text at special angles resulting in garbled text.
TextDiffuser and GlyphControl fail to render tiny-font text and some special characters. 
They all sometimes generate objects of the wrong categories or omit objects.


\label{sec: Ablations}

\begin{table}[!tb]
  \caption{Ablation results of TOF on LTOS dataset.}
  \label{tab: Ablations}
  \centering
  \resizebox{\linewidth}{!}{
  \begin{tabular}{@{}cccccl@{}}
    \toprule
    \textbf{Cross-Attention} & $\mathbf{\mathcal{L}_{text}}$  & $\mathbf{\alpha}$ & \textbf{OCR. ACC} $\uparrow$ & \textbf{NED} $\uparrow$& \textbf{$\text{AP}$} $\uparrow$\\
    \midrule
    $\times$ &$\times$ &$\times$ & 0.3620& 0.5602&0.4512\\
    $\times$ &$\checkmark$ &$\times$ & 0.4014& 0.5951&0.4501\\
    $\checkmark$ &$\checkmark$ &$\times$  & 0.4218&0.6177&0.4662\\
    $\checkmark$ &$\times$  &$\checkmark$ & 0.4268 & 0.6203 & 0.5003\\ 
    $\checkmark$ &$\checkmark$ &$\checkmark$ & \textbf{0.4353}&\textbf{0.6325}&\textbf{0.5658}\\
  \bottomrule
  \end{tabular}
  }
\end{table}



\section{CONCLUSION}
\maketitle
We define a unified task LTOS accurately controlling visual text and object generation and propose the LTOS dataset. We propose a TOF model with a text-object self-adaptive cross-attention fusion module balancing text and object synthesis. Experiments show that TOF obtains SOTA on LTOS and performs well in text rendering and object generations. 

\section{Appendix}
\begin{figure*}[h]
\vspace{-18pt}
  \centering
  \includegraphics[width=\textwidth]{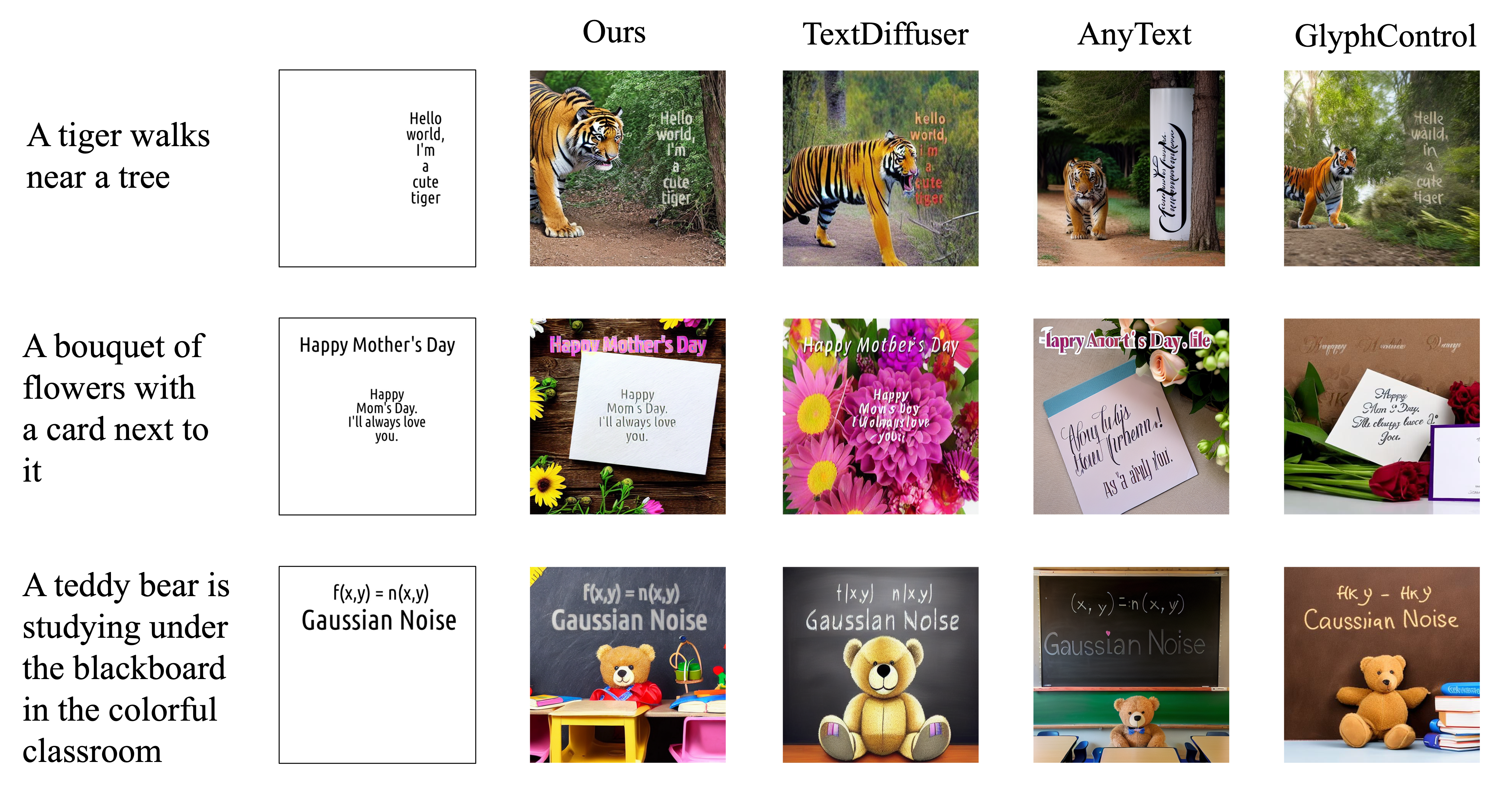}
  \caption{Qualitative results without object layout during inference.
  }
  \label{fig: no_object_layout}
\end{figure*}

\maketitle
\noindent{\textbf{LTOS TASK WITHOUT OBJECT LAYOUT.}} 
In Fig.~\ref{fig: no_object_layout}, we further show several examples generated without the object layout (empty layout map as input for inference). Compared with the baselines, our model maintains a plausible generation ability even after removing the object layout, with clear multi-line text rendered in harmony with the image content.

\begin{figure*}[h]
  \vspace{-10pt}
  \centering
  \includegraphics[width=\textwidth]{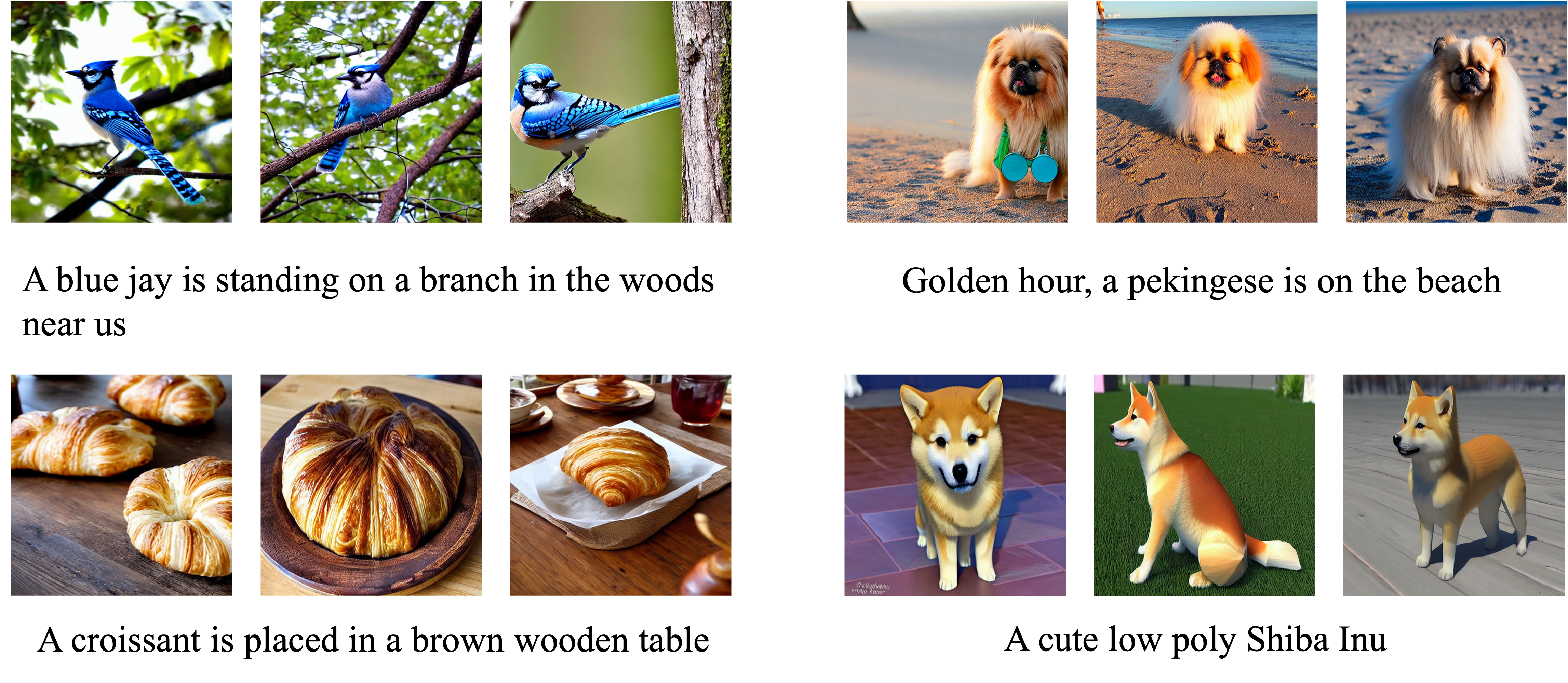}
  \caption{Qualitative results with only text prompt.
  }
  \label{fig: only_prompt}
\end{figure*}

\noindent{\textbf{LTOS TASK WITH ONLY TEXT PROMPT}}
To demonstrate the ability of TOF to generate objects, we generate images without glyph images and object layout maps and show results in Fig.~\ref{fig: only_prompt},. The results show that even without object control, the corresponding objects can be properly generated through only text prompts, such as the blue jay and the brown wooden table. It demonstrates that TOF maintains the ability to generate plausible objects.

\bibliographystyle{IEEEtran}
\bibliography{IEEEfull}


\end{document}